\documentclass[conference]{IEEEtran}
\IEEEoverridecommandlockouts

\usepackage{cite}
\usepackage{amsmath,amssymb,amsfonts}
\usepackage{algorithmic}
\usepackage{graphicx}
\usepackage{textcomp}
\usepackage{xcolor}
\usepackage{amsmath}
\usepackage{multirow}
\usepackage{amsmath}
\usepackage{multirow}
\usepackage{array}
\usepackage{booktabs}
\usepackage{array}
\usepackage{algorithm}
\usepackage{algorithmic}
\usepackage{amsmath}
\usepackage{float}
\usepackage{siunitx}
\usepackage{todonotes}
\usepackage{comment}  
\usepackage{titlesec}
\usepackage{booktabs}
\usepackage{subcaption}
\usepackage{subcaption}
\usepackage{afterpage}
\usepackage{caption}
\usepackage{stfloats} 
\usepackage{color}
\usepackage{placeins}
\usepackage{orcidlink}

\titlespacing*{\paragraph}{0pt}{0pt}{1em}

\def\BibTeX{{\rm B\kern-.05em{\sc i\kern-.025em b}\kern-.08em
    T\kern-.1667em\lower.7ex\hbox{E}\kern-.125emX}}
\makeatletter
\providecommand{\sf@counterlist}{}  
\makeatother

\begin{document}

\title{Hardware-Aware Fine-Tuning of Spiking Q-Networks on the SpiNNaker2 Neuromorphic Platform \\

}





\author{\IEEEauthorblockN{
Sirine Arfa\,\orcidlink{0009-0009-3230-7107}\IEEEauthorrefmark{1}\IEEEauthorrefmark{2},
Bernhard Vogginger\,\orcidlink{0000-0001-9042-5405}\IEEEauthorrefmark{1},
Christian Mayr\,\orcidlink{0000-0003-3502-0872}\IEEEauthorrefmark{1}\IEEEauthorrefmark{6}\IEEEauthorrefmark{4}
}
\IEEEauthorblockA{\IEEEauthorrefmark{1}Chair of Highly-Parallel VLSI-Systems and Neuro-Microelectronics, Technische Universität Dresden, Germany}
\IEEEauthorblockA{\IEEEauthorrefmark{6}ScaDS.AI Dresden/Leipzig, Germany}
\IEEEauthorblockA{\IEEEauthorrefmark{4}Centre for Tactile Internet with Human-in-the-Loop (CeTI)}
\IEEEauthorblockA{\IEEEauthorrefmark{2}Email: sirine.arfa@tu-dresden.de}
}


\maketitle

\begin{abstract}
Spiking Neural Networks (SNNs) promise orders-of-magnitude lower power consumption and low-latency inference on neuromorphic hardware for a wide range of robotic tasks.

In this work, we present an energy-efficient implementation of a reinforcement learning (RL) algorithm using quantized SNNs to solve two classical control tasks. The network is trained using the Q-learning algorithm, then fine-tuned and quantized to low-bit (8-bit) precision for embedded deployment on the SpiNNaker2 neuromorphic chip. To evaluate the comparative advantage of SpiNNaker2 over conventional computing platforms, we analyze inference latency, dynamic power consumption, and energy cost per inference for our SNN models, comparing performance against a GTX 1650 GPU baseline.

Our results demonstrate SpiNNaker2’s strong potential for scalable, low-energy neuromorphic computing, achieving up to 32$\times$ reduction in energy consumption. Inference latency remains on par with GPU-based execution, with improvements observed in certain task settings, reinforcing SpiNNaker2’s viability for real-time neuromorphic control, making the neuromorphic approach a compelling direction for efficient deep Q-learning.
\end{abstract}

\begin{IEEEkeywords}
Spiking neural networks, reinforcement learning, neuromorphic hardware, quantization, energy-efficient computing
\end{IEEEkeywords}

\section{Introduction}
Spiking neural networks (SNNs) have emerged as a promising computational model that combines biological inspiration with practical engineering advantages most notably, sparse activation, event-driven processing, and ultra-low power consumption \cite{maass1997networks}. Unlike conventional artificial neural networks (ANNs), which rely on continuous-valued activations and synchronous computation, SNNs process information through discrete spikes, enabling asynchronous updates \cite{yao2024spike} and significantly reducing computational load. These properties make SNNs an attractive candidate for energy-constrained and latency-sensitive applications \cite{yamazaki2022spiking}, especially when paired with specialized neuromorphic hardware platforms designed to exploit their unique characteristics \cite{nazeer2024language, su2024snn}.

As the demand for on-chip intelligence continues to grow, neuromorphic systems such as SpiNNaker2 \cite{gonzalez2024spinnaker2} have been developed to support scalable and power-efficient execution of SNNs. By embracing asynchronous, parallel architectures that align with the sparse, temporal nature of spiking computation, these platforms offer a powerful  path toward embedded aritificial intelligence (AI) capable of real-time inference without relying on large, power-hungry accelerators \cite{rajendran2019low, davies2018loihi}. While supervised learning with SNNs has seen significant progress thanks in large part to surrogate gradient training,  applications in RL remain relatively underexplored.

RL is a natural match for SNNs \cite{feng2024brainqn}: both are inherently temporal processes that unfold over time \cite{zheng2024temporal}, and the internal state carried by spiking neurons (e.g., membrane potential) offers a lightweight, biologically inspired form of short-term memory. Despite this theoretical alignment, most prior approaches either focus on shallow, biologically motivated algorithms such as reward-modulated spike-timing dependent plasticity (STDP) \cite{kheradpisheh2018stdp}, or rely on ANN-to-SNN conversion techniques \cite{rueckauer2017conversion} that neglect the dynamic structure of spike-based learning during training. While RL with SNNs has shown promise in recent studies \cite{zanatta2024exploring, zanatta2023directly}, practical deployment on embedded neuromorphic platforms remains a complex challenge due to the tight coupling between algorithmic design and hardware constraints.

Recent experimental results have demonstrated that neuromorphic reinforcement learning can deliver substantial efficiency gains across a range of application domains. In robotic control, spiking policies deployed on Intel Loihi achieved real-time performance in force-sensitive manipulation tasks with up to an order-of-magnitude reduction in energy consumption compared to conventional platforms \cite{amaya2023neurorobotic}. 

Similarly, population-coded actor-critic architectures running on neuromorphic chips have been shown to match the performance of standard deep RL methods while consuming over 100× less energy in continuous control settings \cite{tang2021deep}. These advantages stem from the close alignment between neuromorphic hardware and the sparse, temporally structured nature of spike-based computation, enabling event-driven updates and efficient state encoding. As a result, RL on neuromorphic platforms holds particular promise for power-constrained, latency-critical applications such as autonomous robotics, edge AI, and closed-loop control in real-world environments.

In this work, we build upon the model architecture proposed in \cite{akl2021porting}, used SnnTorch \cite{eshraghian2021training} for training the rate-coded SNN models, fine-tuned them for energy-efficient deployment on the SpiNNaker2 neuromorphic platform. Our pipeline involves training spiking agents using the Q-learning algorithm \cite{chen2022deep} in simulation, followed by quantization to 8-bit precision to meet the memory and compute constraints of embedded neuromorphic execution. This approach enables a direct evaluation of the real-world viability of deploying deep spiking RL agents on hardware optimized for low-power operation, while preserving task performance.

This workflow addresses both algorithmic and system-level challenges: enabling deep RL in a spike-based network while ensuring compatibility with energy-efficient hardware execution.

\begin{figure}[t]
    \centering
    \includegraphics[width=0.8\linewidth]{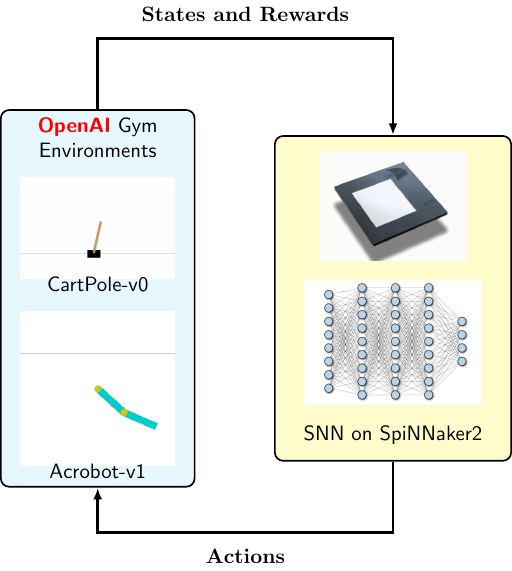}
    \caption{
        \textbf{Pipeline Overview: Closed-loop SNN-based RL with SpiNNaker2.}
        Observations from the OpenAI Gym environments are processed by a quantized spiking neural network deployed on the SpiNNaker2 neuromorphic platform. 
        The system runs in closed-loop, with selected actions sent back to the environment. 
        The SNN architecture includes two fully connected hidden layers; additional details are provided in Table~\ref{table:network_architecture}.
    }
    \label{fig:pipeline_overview}
\end{figure}

The overall system architecture used in our experiments is illustrated in Figure~\ref{fig:pipeline_overview}.
We validate our approach on two classical control benchmarks from the OpenAI Gym \cite{brockman2016openai} suite, focusing not only on cumulative reward but also on practical deployment metrics including dynamic power, latency, and energy per inference. Compared to a conventional GPU baseline (GTX 1650), our system achieves up to a 96\% reduction in energy usage and a 32\% reduction in inference latency, while maintaining competitive task performance. These results demonstrate the viability of neuromorphic hardware as a platform for efficient deep RL and offer a concrete step toward deploying intelligent, learning-enabled agents in constrained environments such as mobile robots, IoT devices, and autonomous vehicles \cite{tang2020reinforcement, tang2021deep}, \cite{hwu2017self}.

\section{Background}

\subsection{Deep Q-Learning}

Q-learning is a model-free, off-policy RL algorithm \cite{mnih2015human} that estimates the optimal action-value function \( Q^*(s, a) \), representing the expected return from taking action \( a \) in state \( s \) and following the optimal policy thereafter:

\begin{equation}
Q^*(s, a) = \max_\pi \mathbb{E} \left[ \sum_{t=0}^\infty \gamma^t r_t \,\bigg|\, s_0 = s,\, a_0 = a,\, \pi \right],
\end{equation}

where \( \gamma \in [0, 1) \) is the discount factor determining the importance of future rewards, and \( r_t \) is the reward at time step \( t \).

Q-values are updated based on observed transitions \( (s, a, r, s') \) using the Bellman equation:

\begin{equation}
Q(s, a) \leftarrow Q(s, a) + \alpha \left[ r + \gamma \max_{a'} Q(s', a') - Q(s, a) \right],
\end{equation}

where \( \alpha \) is the learning rate, \( s' \) denotes the next state resulting from action \( a \), and \( a' \) denotes the action selected in \( s' \).

In high-dimensional environments, Deep Q-Learning approximates \( Q(s, a) \) with a neural network \( Q(s, a; \theta) \), where \( \theta \) are the parameters of the network. To improve training stability, a separate target network with parameters \( \theta^- \) is maintained; these parameters are periodically copied from \( \theta \) and held fixed between updates. Learning proceeds by storing transitions \( (s, a, r, s') \) in a replay memory buffer \( \mathcal{D} \). During optimization, minibatches are sampled uniformly at random from \( \mathcal{D} \), and the parameters \( \theta \) are updated to minimize the temporal difference loss:

\begin{equation}
\mathcal{L}(\theta) = \left( r + \gamma \max_{a'} Q(s', a'; \theta^-) - Q(s, a; \theta) \right)^2.
\end{equation}

This approach enables the agent to learn effective policies directly from high-dimensional inputs and generalize across diverse tasks.

\subsection{The CartPole and Acrobot Tasks}

\subsubsection{CartPole}

The CartPole environment \cite{barto1983neuronlike} involves a pole hinged to a cart that moves along a horizontal track. The system state is given by a 4-dimensional vector \( s \in \mathbb{R}^4 \), comprising the cart's position and velocity, and the pole's angle and angular velocity. The cart can move left and right on a one-dimensional track while the pole is free to move only in the plane vertical to the cart and track, and the agent chooses from a discrete action set $A = \{0, 1\}$, applying a fixed force to either side of the cart.

The objective is to learn a control policy \( \pi: \mathbb{R}^4 \rightarrow \mathcal{A} \) that maximizes cumulative reward by keeping the pole upright. A positive reward is given at each timestep the pole remains within a predefined angular range. In our experiments, an episode terminates when the pole angle exceeds \( \pm \frac{\pi}{12} \) radians or after 200 timesteps.

\subsubsection{Acrobot}

The Acrobot environment \cite{sutton1995generalization} simulates a two-joint pendulum where only the second joint is actuated. The agent receives a 6-dimensional observation \( s \in \mathbb{R}^6 \), consisting of \( \cos\theta_1, \sin\theta_1, \cos\theta_2, \sin\theta_2 \), and the angular velocities \( \dot{\theta}_1, \dot{\theta}_2 \). Actions are chosen from \( \mathcal{A} = \{-1, 0, +1\} \), representing either applying a positive, negative, or
no torque on the joint between the two links. The objective is to swing the lower link upward to raise the tip of the pendulum above a specified vertical threshold as quickly as possible. A constant reward of \(-1\) is given at each step that do not reach the goal, incentivizing the agent to reach the target height promptly. An episode ends when the tip of the pendulum reaches the required height or after 500 steps. The policy \( \pi: \mathbb{R}^6 \rightarrow \mathcal{A} \) is trained to coordinate the motion effectively under limited actuation.

\section{Methodology}

\subsection{Model Architecture}
The network architecture and simulation parameters used in our experiments are summarized in Table~\ref{table:network_architecture}, while the DQN training hyperparameters are provided in Table~\ref{table:dqn_hyperparams}. The Deep Q-learning algorithm from \cite{akl2021porting} was fine-tuned and used to train the spiking agent, leveraging backpropagation \cite{rumelhart1986learning} through time with surrogate gradients, as implemented via the snnTorch framework~\cite{eshraghian2021training}. The model consists of fully connected layers interleaved with leaky integrate-and-fire (LIF) neurons \cite{eshraghian2022navigating}. Each neuron's membrane potential evolves over time according to:

\begin{equation}
u^{j}_{t+1} = \beta u^{j}_{t} + \sum_i w^{ij} z^{i}_{t} - z^{j}_{t} \theta,
\label{eq:lif_dynamics}
\end{equation}

where \( u^j_t \) is the membrane potential of neuron \( j \) at time \( t \), \( \beta \in [0,1) \) is the decay factor, \( w^{ij} \) are the synaptic weights, and \( \theta \) is the firing threshold. A spike \( z^{j}_{t} \in \{0, 1\} \) is emitted when the membrane potential crosses the threshold, and the potential is reset by subtracting \( \theta \).

The input dimensionality of each network is determined by doubling the size of the environment’s observation space. This expansion accounts for the two-neuron signed input encoding scheme, which will be described in detail in Section~\ref{subsec:Inputencoding}. Input states are converted into spike trains using rate coding and propagated through the network over a fixed simulation window. 

Final Q-values are derived from the last-step membrane potentials of the output layer, which does not emit spikes, enabling continuous-valued predictions suitable for gradient-based learning. The network state is reset after every environment interaction to maintain consistency with the original DQN formulation. Training is performed using a mean squared error loss between the predicted Q-values and the target Q-values computed via temporal difference learning.

\begin{table}[ht]
\centering
\caption{DQN algorithm hyperparameters}
\label{table:dqn_hyperparams}
\begin{tabular}{ll}
\toprule
\textbf{Hyperparameter} & \textbf{Value} \\
\midrule
replay memory size              & 10000 \\
discount factor                 & 0.999 \\
learning rate                   & 0.001 \\
target network update frequency & 10 \\
initial exploration             & 0.5 \\
final exploration               & 0.05 \\
batch size                      & 128 \\
\bottomrule
\end{tabular}
\end{table}

\vspace{0.5cm}


\begin{table}[ht]
\centering
\caption{SNN parameters}
\label{table:network_architecture}
\begin{tabular}{lcc}
\toprule
\textbf{Hyperparameter} & \textbf{Cartpole-v0} & \textbf{Acrobot-v1} \\
\midrule
$\beta$                 & 0.8   & 0.8   \\
threshold               & 0.5 & 0.5 \\
Reset Mechanism         & Subtract & Subtract \\
Delay         & 0 & 0 \\
simulation time         & 10  & 20   \\
network architecture    & [8, 64, 64, 2] & [12, 256, 256, 3] \\
\bottomrule
\end{tabular}
\end{table}

\subsection{Model Quantization}
\label{subsec:quantization}

To prepare the trained deep spiking Q-network (DSQN) models for deployment on SpiNNaker2 hardware, we quantize all synaptic weights to 8-bit signed integers in accordance with hardware constraints supporting values in the dynamic range of $[-128,127]$~\cite{arfa2025efficient}. After training in full precision, we extract the weight tensors from each fully connected layer and apply a scaling and clipping procedure to map floating-point values to 8-bit integers. Specifically, weights are multiplied by a fixed scaling factor $\lambda$ and normalized by the maximum absolute value to utilize the integer range, followed by rounding and type casting. Formally, each floating-point weight $w \in \mathbb{R}$ is mapped to an 8-bit signed integer $\hat{w} \in \mathbb{Z}$ as:

\begin{equation}
\hat{w} = \left\lfloor w \cdot \lambda \cdot \underbrace{\frac{127}{\max(|w|)}}_{\lambda_s} \right\rfloor,
\end{equation}

where $\lambda$ adjusts the effective spread of the weight distribution, and $\lambda_s$ scales the maximum weight to 127. While $\lambda_s$ ensures the largest value reaches the upper bound, it does not affect the variance of smaller weights, which often cluster near zero in RL models. Applying $\lambda > 1$ expands this range, preserving resolution and preventing quantization collapse, at the cost of increased clipping risk. Empirically, more complex tasks (e.g., Acrobot-v1) required higher scaling factors ($\lambda=32$), while simpler tasks (e.g., CartPole-v0) performed well with moderate scaling ($\lambda=3$). The model is trained offline in full precision, and quantization is applied post-training to enable on-chip inference using 8-bit weights.


\begin{figure*}[!t]
    \centering
    \subfloat[CartPole-v0]{%
        \includegraphics[width=0.95\textwidth]{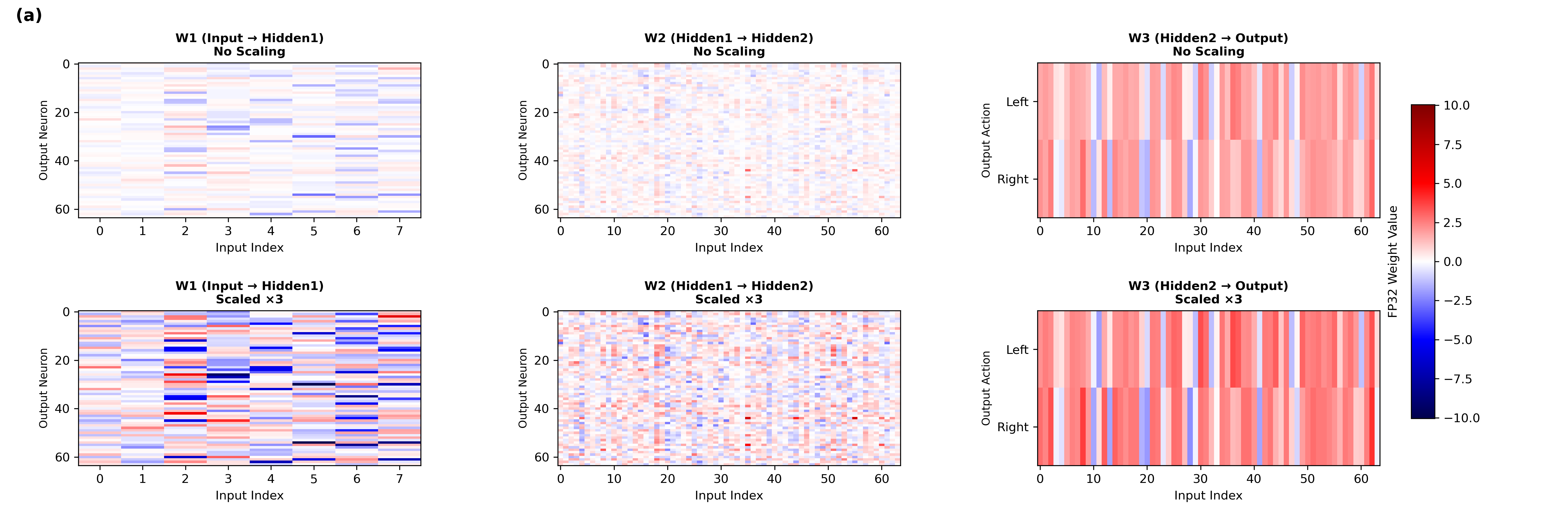}
    }\par\vspace{2mm}
    \subfloat[Acrobot-v1]{%
        \includegraphics[width=0.95\textwidth]{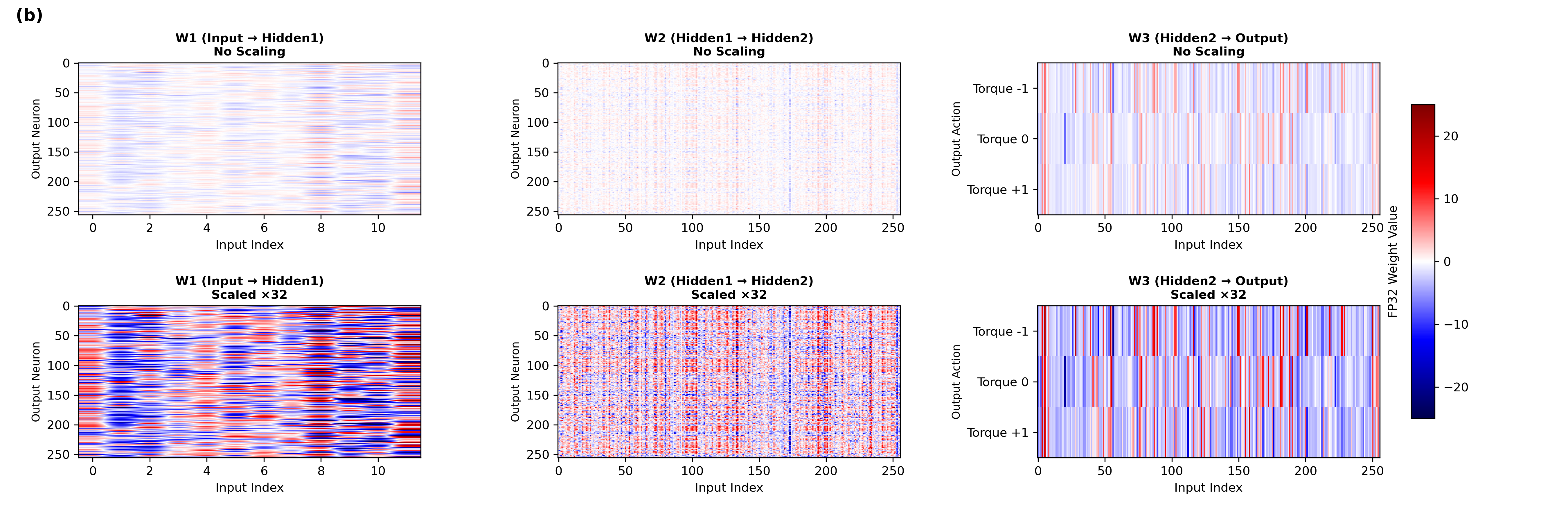}
    }
    \caption{
        \textbf{Effect of Uniform Full-Layer Quantization Scaling in CartPole-v0 and Acrobot-v1.}
        (a) In CartPole-v0, applying a uniform scaling factor across all layers ($\times 3$) increases the dynamic range of early-layer weights without saturating the network, preserving temporal sparsity and enabling maximum reward performance.
        (b) For Acrobot-v1, a more challenging control task, full-layer scaling ($\times 32$) is essential to expand hidden-layer dynamics. Without it, weights collapse near zero, limiting signal propagation and reducing task performance.
    }
    \label{fig:activity_comparison}
\end{figure*}

\begin{figure*}[!t]
    \centering
    \subfloat[CartPole-v0]{%
        \includegraphics[height=6cm]{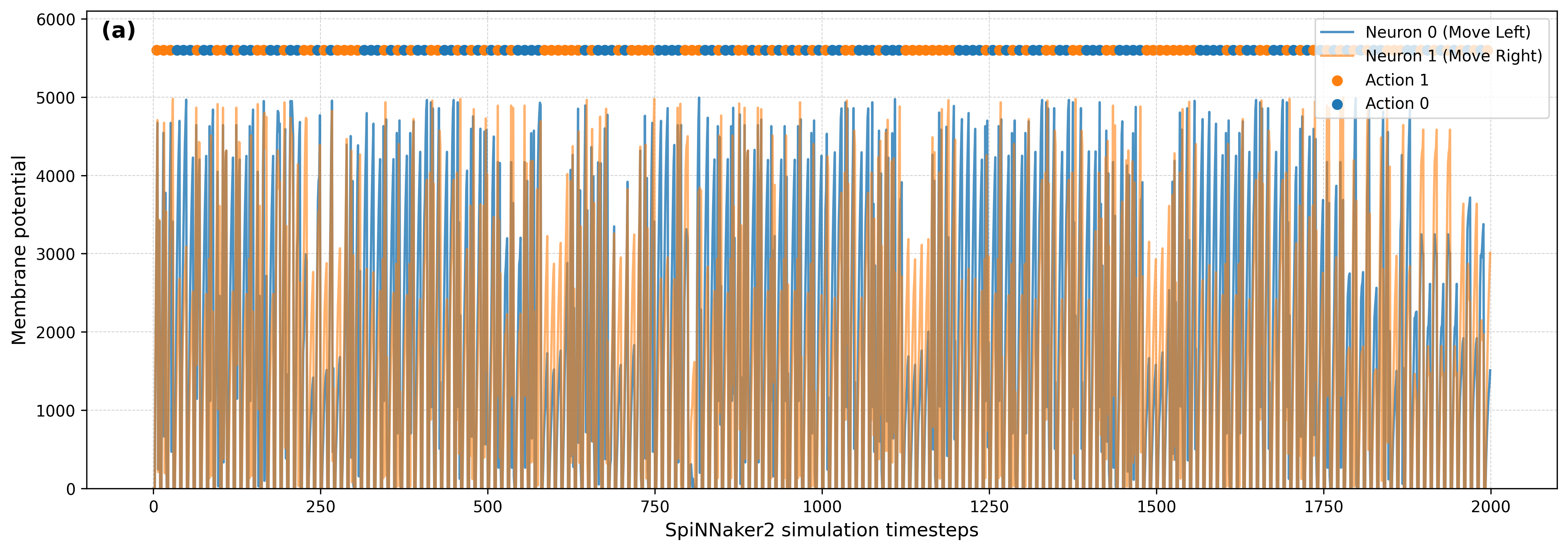}
    }\hfill
    \subfloat[Acrobot-v1]{%
        \includegraphics[height=6cm]{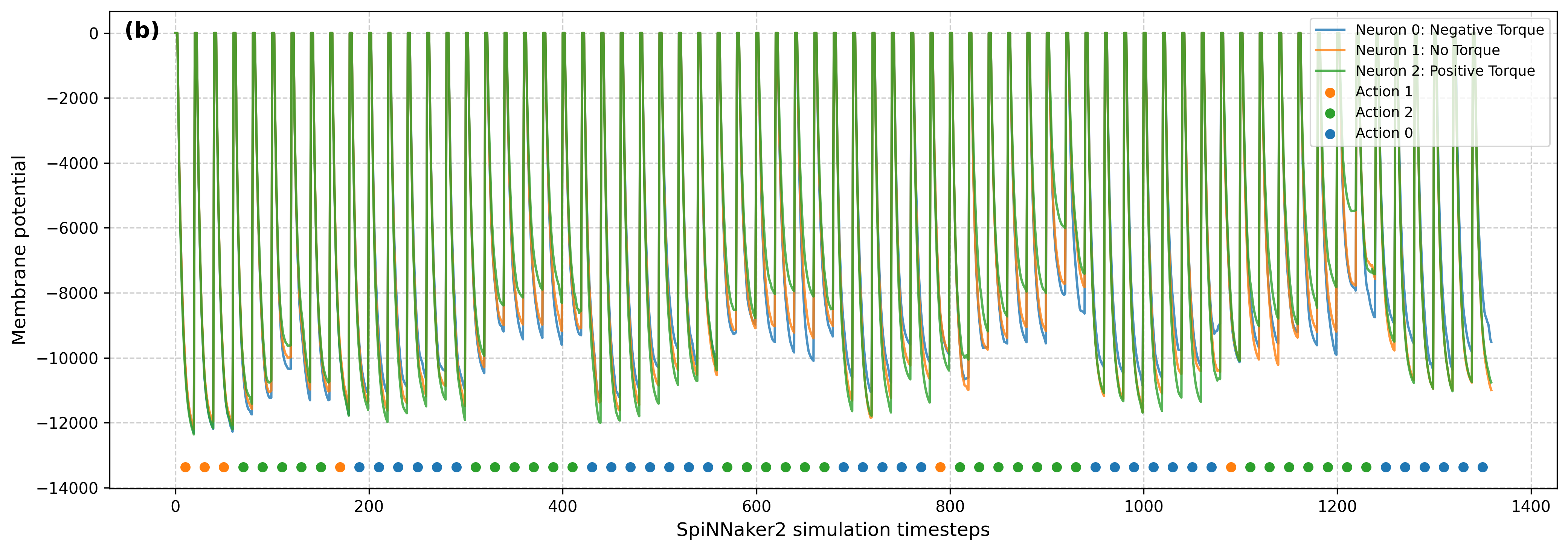}
    }
    \caption{
        \textbf{Voltage dynamics in CartPole-v0 and Acrobot-v1 recorded for one episode on SpiNNaker2.}
        Horizontal axis: the number of time steps of the episode multiplied by the simulation time of each trial, listed in Table~\ref{table:network_architecture}. The dots indicate which action was selected at each time step, based on the neuron with the higher membrane potential. In the example runs shown here, the CartPole-v0 episode lasted 200 time steps (reward = 200); the Acrobot-v1 episode lasted 67 time steps (reward = -67).
    }
    \label{fig:V_comparison}
\end{figure*}

\subsection{Encoding and Decoding Strategies}
\label{subsec:Inputencoding}

\paragraph{Input Encoding}
To represent continuous-valued input states as spike-based signals compatible with neuromorphic hardware, we adopt a two-neruon encoding scheme similar to the encoding used in \cite{akl2023toward}. First, each input feature is transformed using a signed two-neuron encoding: positive values are assigned to one neuron and negative values to another, yielding a doubled input dimensionality. Formally, a scalar value $x \in \mathbb{R}$ is represented as the vector $(x, 0)$ if $x \geq 0$, and $(0, |x|)$ otherwise.

The resulting vector is then encoded using rate coding, where the firing rate of each input neuron is proportional to the encoded value over a fixed simulation window. The spike-based representation is generated using poisson-distributed spike trains sampled over $T$ discrete timesteps. These are then converted into a spike list format, specifying the exact firing times of each neuron, in order to be compatible with SpiNNaker2’s spike list-based input interface.

\paragraph{Output Decoding}

Action selection is performed by reading out the final membrane potentials of the output layer neurons after the full simulation run. To enable continuous-valued outputs while preserving spiking activity in the rest of the network, we configure the output neurons with a high firing threshold ($\theta=5000$), effectively disabling spike emission. Neuron variables, including the threshold, are represented as floats (single precision). This effectively disables spiking in the output layer, allowing the membrane potentials to accumulate over time without being reset. A similar decoding approach using infinite thresholds to suppress spiking has been used in~\cite{akl2021porting}. In our case, the elevated threshold ensures that output voltages reflect the integrated response to incoming spikes. The selected action $a^*$ corresponds to the output neuron with the highest voltage at the final timestep:
\[
a^* = \arg \max_j v_j(T),
\]
where $v_j(T)$ denotes the membrane potential of output neuron $j$ at the final timestep $T$. Unlike discrete action problems that rely on spike counts, continuous control tasks benefit from using the final voltage as a precise analog readout without decay or reset, as discussed in~\cite{akl2023toward}.


\section{Results and Evaluation}

\subsection{Quantization Strategy Ablation}

\label{subsec:quantization_ablation}

We evaluated the impact of uniform layer-wise quantization scaling on the performance of quantized DSQNs in the two RL tasks at hand: CartPole-v0 and Acrobot-v1. In CartPole-v0, applying a moderate uniform scaling factor across all layers ($\lambda=3$) successfully preserved spike-based signal propagation and temporal sparsity, while achieving the maximum possible reward of 200, comparable to the performance reported in~\cite{akl2021porting}. For the more complex Acrobot-v1 task, which involves a higher-dimensional state space and underactuated dynamics, a larger scaling factor ($\lambda=32$) was necessary to maintain expressiveness and enable effective policy learning, achieving a maximum reward of $-67$, slightly better than the results reported for the Loihi implementation in~\cite{akl2021porting}. These values were selected empirically after extensive experiments measuring task performance across a range of scaling factors, where the highest average rewards were obtained at $\lambda=3$ for CartPole-v0 and $\lambda=32$ for Acrobot-v1. As visualized in Figure~\ref{fig:activity_comparison}, full-layer scaling expands the dynamic range of weights in each layer, particularly in hidden-to-hidden connections, without causing saturation. In future work, we plan to quantify the minimum viable scaling factor required to preserve performance as a function of task complexity, model architecture, and training dynamics.

\begin{figure}[!t]
    \centering
    \subfloat[CartPole-v0]{%
        \includegraphics[width=0.8\linewidth]{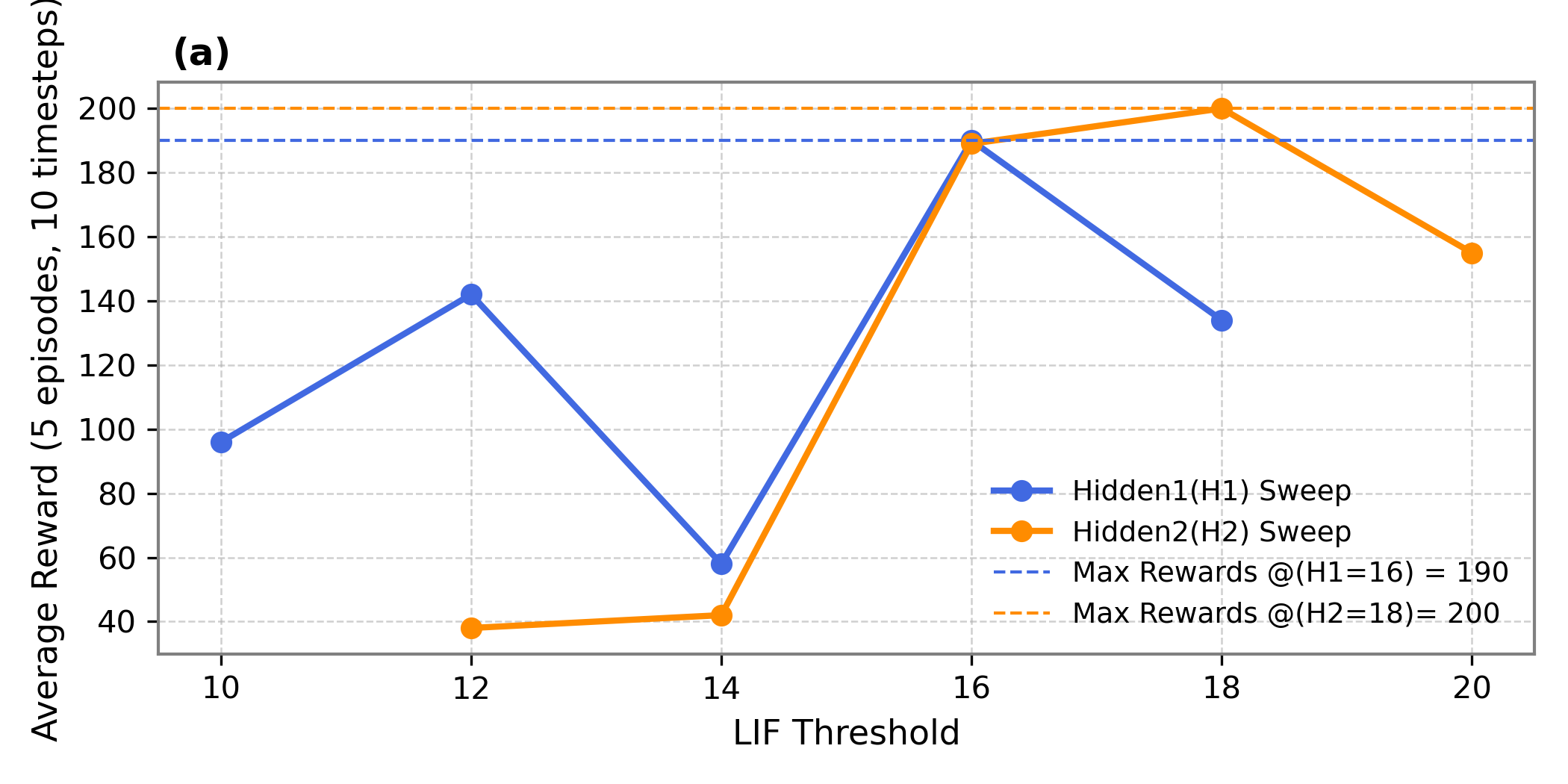}
    }\hfill
    \subfloat[Acrobot-v1]{%
        \includegraphics[width=0.8\linewidth]{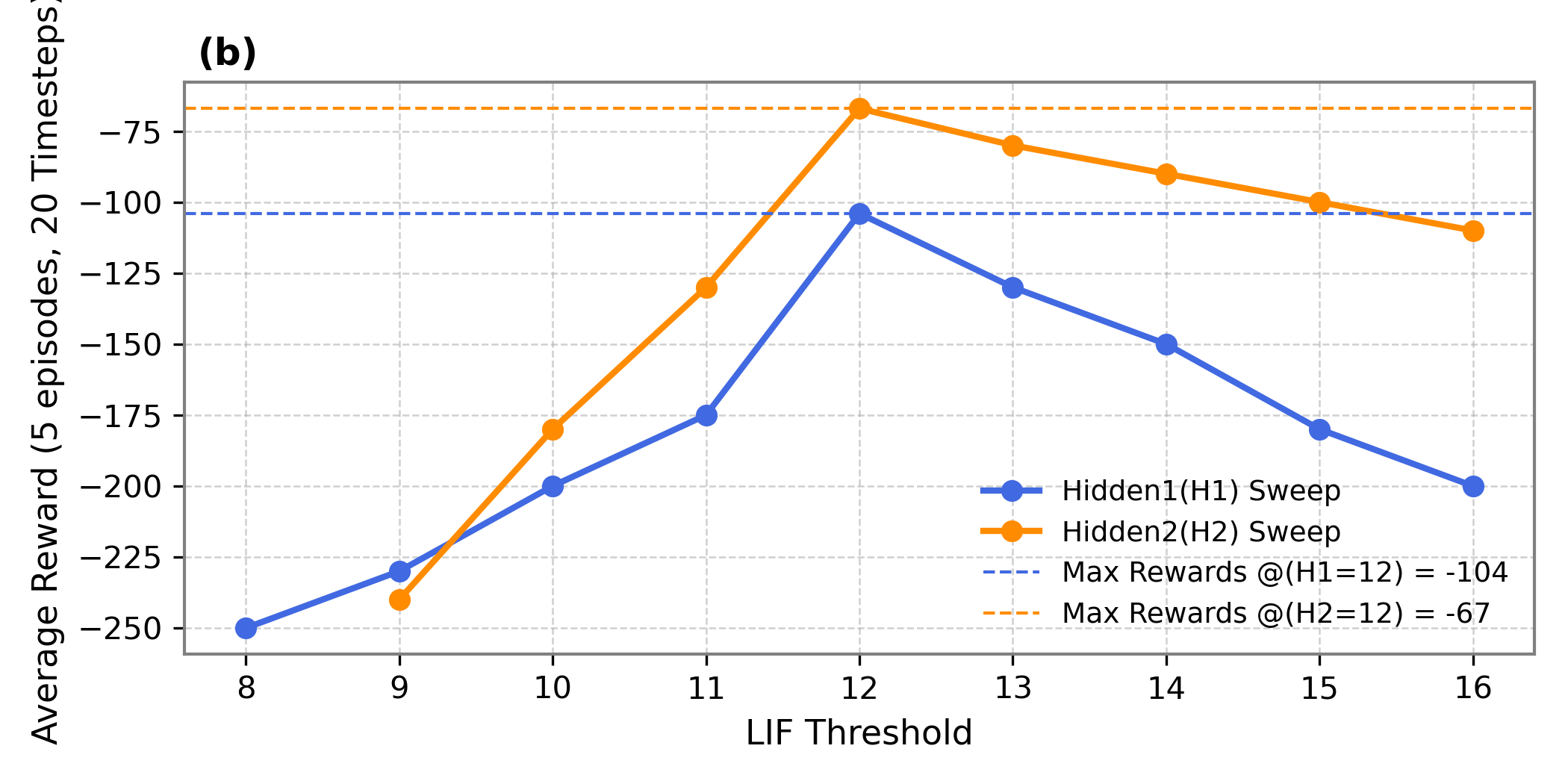}
    }

    \begin{minipage}{\linewidth}
        \caption{
            Average rewards over 5 episodes for various hidden layer thresholds, showing performance sensitivity across tasks.
        }
        \label{fig:eval_barplots}
    \end{minipage}
\end{figure}

\begin{figure*}[!t]
    \centering
    \subfloat[CartPole-v0]{%
        \begin{minipage}[b]{0.31\textwidth}
            \includegraphics[width=\textwidth]{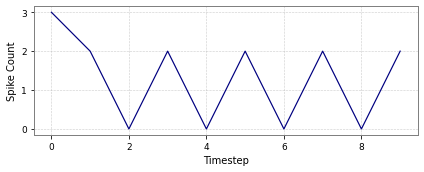}\\[1mm]
            \includegraphics[width=\textwidth]{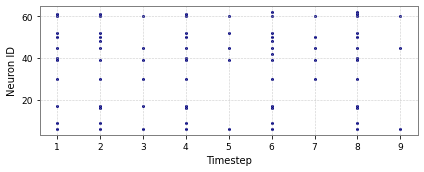}\\[1mm]
            \includegraphics[width=\textwidth]{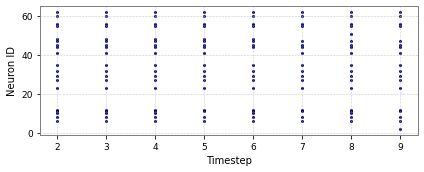}
        \end{minipage}
    }\hspace{10mm}
    \subfloat[Acrobot-v1]{%
        \begin{minipage}[b]{0.31\textwidth}
            \includegraphics[width=\textwidth]{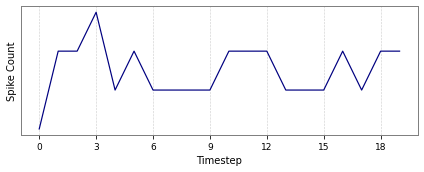}\\[1mm]
            \includegraphics[width=\textwidth]{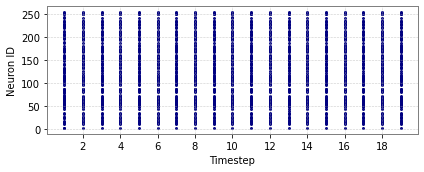}\\[1mm]
            \includegraphics[width=\textwidth]{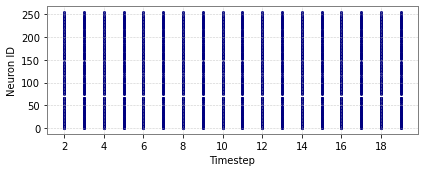}
        \end{minipage}
    }
    \caption{
        \textbf{Spike activity for CartPole-v0 and Acrobot-v1 across input and hidden layers.}
        Each column corresponds to a task (CartPole-v0 on the left, Acrobot-v1 on the right), and each row displays spike activity for a layer (input, hidden 1, hidden 2). The first row shows spike counts over time, and the second and third rows present spike raster plots across neurons. These visualizations highlight differences in temporal dynamics and task complexity between the two environments.
    }
    \label{fig:spike_activity}
\end{figure*}

\subsection{On-Chip Threshold Sensitivity Analysis}

To determine optimal LIF thresholds for on-chip deployment, we performed a two-stage grid search over the hidden layers of the trained network. In the first stage, we fixed the threshold of the second hidden layer to a random value and swept over a range of thresholds for the first hidden layer, measuring the average reward over 5 evaluation episodes per setting. The threshold yielding the highest average reward was selected. In the second stage, we fixed this optimal threshold for the first hidden layer and repeated the sweep for the second hidden layer. This sequential tuning procedure enabled us to efficiently identify high-performing threshold configurations while reducing the combinatorial cost of full joint optimization. Because quantization substantially alters weight magnitudes and distributions, thresholds cannot be deterministically rescaled from simulation values. We therefore sweep thresholds empirically to re-balance neuron excitability and recover the target spiking dynamics. The results, shown in Figure~\ref{fig:eval_barplots}, highlight the sensitivity of DSQN performance to threshold values, underscoring the importance of careful hardware-aware tuning.

\subsection{Meeting Real-Time Constraints with SpiNNaker2 for Closed-Loop Control}

After training two Deep DSQN models to solve the CartPole-v0 and Acrobot-v1 tasks, we quantized the models as described in Section~\ref{subsec:quantization} and deployed them on the SpiNNaker2 neuromorphic chip. A consistent finding across both environments is that across both tasks, we observe no degradation in policy performance when deploying the trained DSQN models on SpiNNaker2. In fact, the neuromorphic hardware achieves performance equal to or slightly better than the software simulator baseline (snnTorch), with Acrobot-v1 exhibiting a modest improvement in total reward on-chip as shown in Table~\ref{tab:metric_summary}.

For the deployment on the chip we use the software py-spinnaker2 \cite{vogginger2024pyspinnaker2} that provides a light-weight Python interface for running experiments on a single-chip SpiNNaker2 test board. It uses 8-bit signed synapse weights and 32-bit floating-point
numbers for neuron parameters and state variables respectively. The API for defining SNN models is inspired by pyNN \cite{davison2009pynn}.

Figure~\ref{fig:V_comparison} shows an exemplary episode from each environment recorded on SpiNNaker2, illustrating membrane potential dynamics in the output layer along with the corresponding selected actions. In both cases, the voltage traces reveal task-aligned decision behavior: CartPole-v0 exhibits sparse, stable membrane activity with well-defined action transitions, while Acrobot-v1 displays denser, oscillatory patterns reflecting its more complex control dynamics.

Note that the output membrane potential can accumulate positive or negative values depending on the relative activity of hidden neurons and the sign distribution of their output weights. While the output weights contain both positive and negative entries, the environment-specific input dynamics lead to different spiking patterns. In CartPole, neurons associated with positive output weights are more active on average, causing positive voltage accumulation. In contrast, Acrobot produces spiking activity patterns that more strongly excite neurons with negative output weights, leading to negative membrane potentials. Since no reset or decay is applied in the output layer, these contributions accumulate over the full episode duration.

Further evidence of proper neuromorphic execution is shown in the spike activity patterns in Figure~\ref{fig:spike_activity}. In CartPole-v0, spiking remains sparse and localized across layers, reflecting efficient inference and temporal sparsity. In contrast, Acrobot-v1 demonstrates significantly higher spike rates and more distributed activation, consistent with the need for richer internal dynamics to solve underactuated and delayed-reward problems.

Together, these results confirm that the quantized DSQN models maintain both functional performance and biologically inspired temporal coding during closed-loop inference on SpiNNaker2.

To ensure real-time interaction between SpiNNaker2 and the OpenAI Gym environments, we enforced a consistent wall-clock control interval mirroring the simulated processing time on hardware. SpiNNaker2 operates with a tick-based scheduler where each tick corresponds to 1\,ms ($T_{\text{tick}}=1.0 \times 10^{-3}\,\mathrm{s}$). We defined per-step simulation durations $T_{\text{sim}}$ of 20 ticks for Acrobot-v1 and 10 ticks for CartPole-v0, aligned with their control frequencies (50\,Hz and 100\,Hz, respectively). The timing logic was:
\[
T_{\text{ctrl}} = T_{\text{sim}} \times T_{\text{tick}}, \quad T_{\text{delay}} = \max(0, T_{\text{ctrl}} - T_{\text{inf}})
\]
where $T_{\text{inf}}$ is the measured wall-clock duration of inference, and $T_{\text{delay}}$ is the artificial delay added to maintain real-time pacing.

This strategy issued actions at biologically plausible intervals rather than as fast as possible. Critically, introducing this timing control did not degrade performance: in Acrobot-v1, the policy maintained an average reward of $-67$, and in CartPole-v0, the agent achieved the maximum reward of $200$, validating SpiNNaker2's suitability for real-time closed-loop control.

\begin{table*}[t]
    \centering
    \caption{Comparison of latency, energy consumption, and power usage between SpiNNaker2 and Nvidia GPU GTX 1650 platforms for one episode of Cartpole-v0 and Acrobot-v1 tasks.}
    \label{tab:metric_summary}
    \renewcommand{\arraystretch}{1.3}
    \begin{tabular}{lcccc}
        \toprule
        \textbf{Measurement} & \multicolumn{2}{c}{\textbf{SpiNNaker2}} & \multicolumn{2}{c}{\textbf{GTX 1650}} \\
        \cmidrule(lr){2-3} \cmidrule(lr){4-5}
        & Cartpole-v0 & Acrobot-v1 & Cartpole-v0 & Acrobot-v1 \\
        \midrule
        Power [W]      & 0.335 & 0.333 & 7.39 & 7.29 \\
        Energy [J]     & 0.006 & 0.010 & 0.145 & 0.321 \\
        Duration [s]   & 0.020 & 0.030 & 0.0197 & 0.0440 \\
        Rewards        & 200 & -67 & 200 & -75 \\
        \bottomrule
    \end{tabular}
\end{table*}

\subsection{Comparison with GPU inference}

Table~\ref{tab:metric_summary} reports a comparison of energy consumption, execution time, and power draw between SpiNNaker2 and an Nvidia GTX 1650 GPU. Across both CartPole-v0 and Acrobot-v1, SpiNNaker2 demonstrates substantial energy efficiency improvements. For example, completing a CartPole-v0 episode requires 0.006~J on SpiNNaker2 versus 0.145~J on the GPU, a 24$\times$ reduction. Similarly, Acrobot-v1 consumes 0.010~J compared to 0.321~J on the GPU, achieving a 32$\times$ reduction. Inference latency remains comparable, but SpiNNaker2 operates with an average power draw below 0.34~W, far less than the GPU’s over 7~W. These results highlight SpiNNaker2 as a promising low-power platform for real-time neuromorphic reinforcement learning in edge and embedded applications.

\section{Discussion and Conclusion}

In this work, we demonstrated the integration of deep reinforcement learning with SNNs, using surrogate gradient training to solve the CartPole-v0 and Acrobot-v1 control tasks from OpenAI Gym. After quantizing the trained DSQN models, we deployed them on the SpiNNaker2 neuromorphic chip and evaluated them in closed-loop interaction with the environments.

A central contribution is the hardware-aware fine-tuning process, which enabled lossless deployment. We performed a two-stage grid search to optimize LIF thresholds and applied uniform layer-wise scaling to preserve representational capacity after quantization. These strategies were critical to maintaining robust spike-based signal propagation, especially in Acrobot-v1. The results show that calibrated quantization and threshold tuning preserved both accuracy and biologically inspired temporal dynamics on hardware.

Compared to GPU-based inference, our neuromorphic implementation achieved up to 32$\times$ higher energy efficiency with minimal latency overhead. Membrane potential traces and spike activity visualizations confirm that the SNNs exhibit task-aligned behavior when deployed on SpiNNaker2, validating both functional accuracy and hardware efficiency.

This work paves the way for broader use of neuromorphic systems in real-time reinforcement learning, especially for energy-constrained edge and robotics scenarios. Beyond inference, our findings encourage exploring on-chip or in-the-loop training with event-driven computation. Integrating learning and control on low-power platforms like SpiNNaker2 represents a promising direction for scaling intelligent agents in real-world applications.

We provide the code used for all experiments at:

\url{https://gitlab.com/Sirine_Arfa/interactive-reinforcement-learning-on-spinnaker2.git}

\section*{Acknowledgment} 

The authors thank Johannes Partzsch for his valuable feedback on the manuscript. This work is funded by the European Union within the programme Horizon Europe under grant agreement no.~101120727 (PRIMI).
Christian Mayr is funded by the German Research Foundation (DFG, Deutsche Forschungsgemeinschaft) as
part of Germany’s Excellence Strategy – EXC 2050/1 – Project ID 390696704 – Cluster of Excellence “Centre for Tactile
Internet with Human-in-the-Loop” (CeTI) of Technische Universit\"{a}t Dresden.

\bibliographystyle{IEEEtran}
\bibliography{references}

\end{document}